\documentclass[onecolumn]{IEEEtran}

\usepackage{graphicx}
\usepackage{multirow}

\begin{document}

\title{Scaling Deep Learning Research with Kubernetes on the NRP Nautilus HyperCluster}

\author{
J. Alex Hurt, Anes Ouadou, Mariam Alshehri, Grant J. Scott
\\
University of Missouri - Columbia
}

\maketitle

\begin{abstract}
  Throughout the scientific computing space, deep learning algorithms have shown excellent performance in a wide range of applications.
  As these deep neural networks (DNNs)
  continue to mature, the necessary compute required to train them has continued to grow.
  Today, modern DNNs require millions of FLOPs and days to weeks of training to generate a well-trained model.
  The training times required for DNNs are oftentimes a bottleneck in DNN research for a variety of deep learning applications, and as such, accelerating and scaling DNN training enables more robust and accelerated research.
  To that end, in this work, we explore utilizing the NRP Nautilus HyperCluster to automate and scale deep learning model training for three separate applications of DNNs, including overhead object detection, burned area segmentation, and deforestation detection.
  In total, 234 deep neural models are trained on Nautilus, for a total time of 4,040 hours.
\end{abstract}

\section{Introduction}

Deep convolutional neural networks (DCNNs) have been established as the state of the art in computer vision (CV) and have shown superior performance in visual tasks for many domains, including remote sensing.
With billions of pixels being collected by overhead sources like satellites, remote sensing (RS) is becoming evermore a big-data problem domain, with endless amounts of data available to enable CV applications.
Due in part to this data availability, the training and optimization of deep networks for RS applications has been explored to great lengths in recent years.
In 2017, researchers investigated utilizing DCNNs for land-cover classification in overhead imagery along with techniques such as transfer learning and data augmentation\cite{CGINets}.
This work was then extended into multi-network fusion research, where multiple DCNNs trained on overhead satellite imagery were fused using simple fusion techniques such as voting and arrogance \cite{GRSL_Deep_Fusion} and then compared to more complex fusion algorithms such as the Choquet and Sugeno Fuzzy Integral \cite{grsl_enhanced_fusion,FUSION_CHAPTER}.

While these studies explored utilizing DCNNs to perform classification on overhead RS imagery, further exploration was required in broad area search, in which DCNNs are trained and used not on clean pre-processed datasets, but instead applied to large swaths of overhead imagery with the goal of finding all instances of a given object or terrain.
In these applications, large swaths of imagery are investigated, usually on the order of tens or hundreds of thousands of square kilometers, if not more.
For example, in \cite{CGI_SAM_SITES}, researchers utilized a DCNN to locate surface-to-air missile sites in the South China Sea.
Surface-to-air missile site detection was then improved in \cite{cannaday2019improved}, in which researchers employed spatial fusion of component detections.
In another application, broad area search was also explored in \cite{angryshift}, in which researchers utilized DCNNs to search for aircraft and other classes in a 9,500 square kilometer AOI surrounding Beijing.
In each of these applications of DCNNs for Broad Area Search, the compute required is a limiting factor in applying these algorithms at scale.

Aside from classification and broad area search, DCNNs have also been utilized to perform object detection on overhead RS imagery.
In \cite{hurt2019comparison}, the YOLOv3 architecture was compared with other deep network architectures to detect military vehicle groups, and in \cite{hurt2020maneuverability}, researchers used overhead imagery to perform maneuverability hazard detection with multiple DCNN-based object detection methodologies.
More recently, as DCNNs have evolved and their operational characteristics have become more clear, researchers have investigated utilizing various techniques to address the shortcomings of DCNNs, including utilizing shape extraction to increase model performance, as in \cite{dmpnet,igarss_dmpnet_2021}.

Throughout CV and RS research, DCNNs are being utilized to enable more real-world applications, however
as DCNNs have matured and the number of convolutional layers has grown, so too have the compute requirements to train models.
Additionally, as DCNN have grown in popularity, several frameworks have been created to enable better scaling and reproducibility of training deep networks.
Among the benefits of these frameworks is the parallelization capabilities built into the training processes, however, these benefits can only be utilized if the appropriate hardware is available.
Large numbers of GPUs are required to effectively train networks in a reasonable amount of time and keep batch sizes large enough to minimize the impact of diminishing gradients.
Further, CPU parallelization is often used for managing GPU resources in distributed training paradigms as well as for asynchronous data loading and preprocessing during the training process.

In this work, we explore utilizing the National Research Platform (NRP) Nautilus HyperCluster to accelerate and scale three separate scientific applications utilizing DCNNs.
Nautilus is an ever-growing Kubernetes cluster containing over 1300 NVIDIA GPUs and 19,000 CPU Cores that can be utilized for a variety of applications, including teaching, research prototyping and exploration, and scaling of scientific compute, which we will explore here.
The three DCNN applications discussed in this study are all in the overhead RS domain: overhead object detection with visual transformer architectures, burned area segmentation with U-Net, and deforestation detection in the Brazilian Amazon rain forest.
Each of these applications requires an intense amount of compute to efficiently and exhaustively study, thus demonstrating the need for the Nautilus HyperCluster.
In total, the research performed on Nautilus for these applications so far totals over 3,000 GPU hours of compute and nearly 250 trained deep neural models.

\section{Scientific Applications}
The compute available in the Nautilus HyperCluster enables a number of applications across the scientific community in a variety of fields.
Herein, we present three such applications requiring the vast compute available in Nautilus in the RS space, all of which utilize DCNNs as their primary methodology.
The first research scaling enabled by Nautilus is an investigation into how visual transformers, which have seen a rise in popularity in ground-based CV research, translate into the RS space.
The next application discussed is wildfire-burned area mapping, in which semantic segmentation models are utilized to build models that can assist with the recovery efforts from wildfires in North America.
Finally, Nautilus is used to accelerate the research into detecting and mapping deforestation of the Brazilian Amazon rainforest.
More details of these domains and their respective methodologies are given in the sections below.

\subsection{Overhead Object Detection with Transformers}
\label{sec:object-detection-methods}

\begin{figure}[tb]
  \centering
  \begin{tabular}{c|c|c}
    \includegraphics[width=0.3\linewidth]{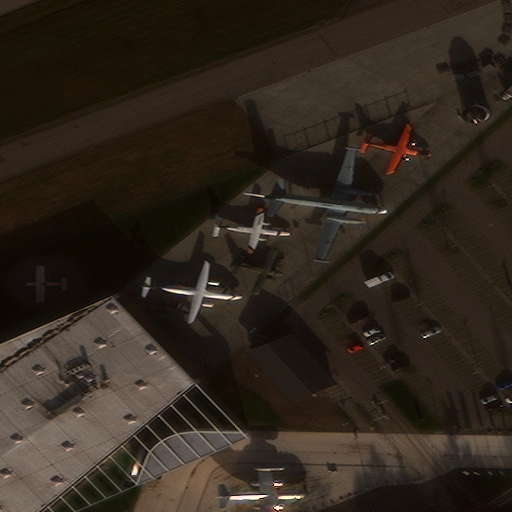} & \includegraphics[width=0.3\linewidth]{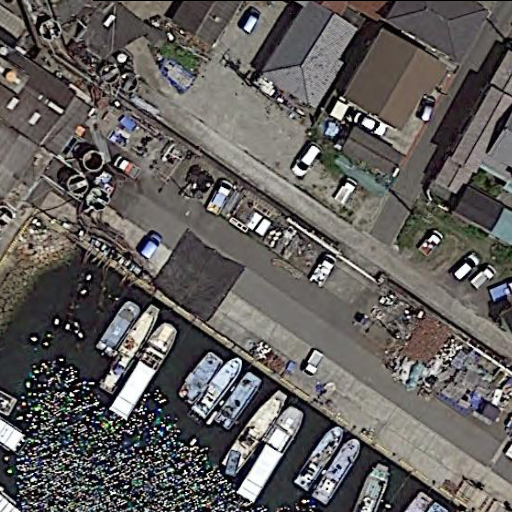} & \includegraphics[width=0.3\linewidth]{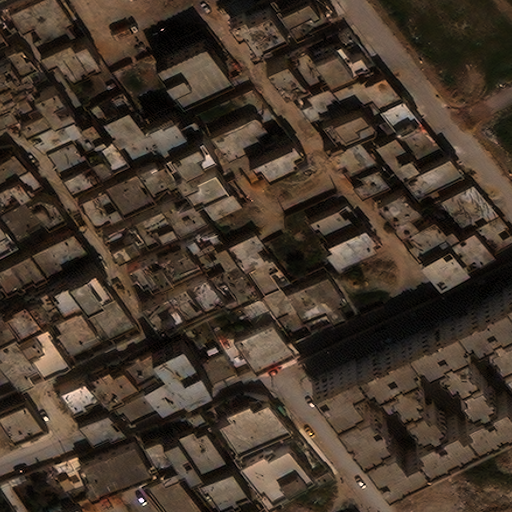} \\
    \includegraphics[width=0.3\linewidth]{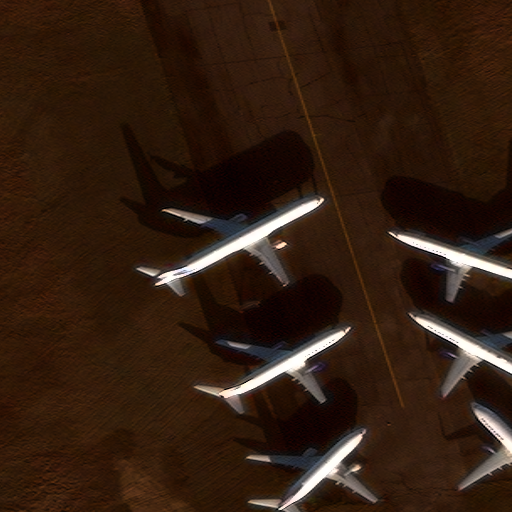} & \includegraphics[width=0.3\linewidth]{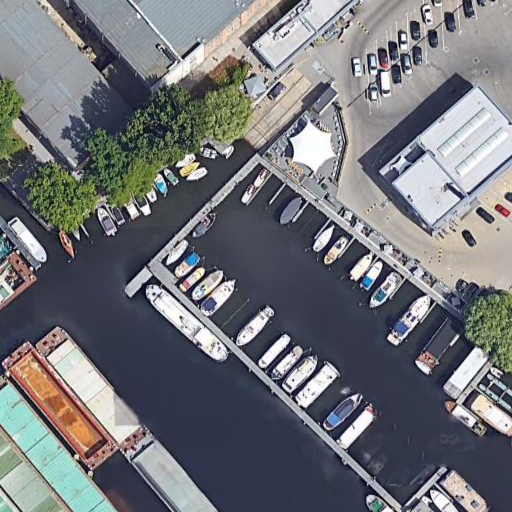} & \includegraphics[width=0.3\linewidth]{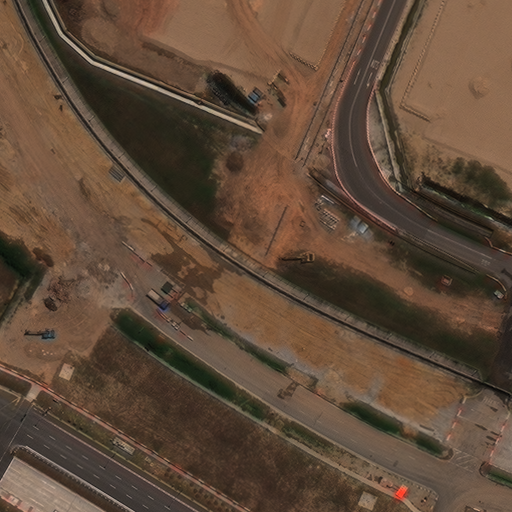} \\
    \hline
    \textbf{RarePlanes}                                                                        & \textbf{DOTA}                                                   & \textbf{XView}                                                  \\
  \end{tabular}
  \caption{Sample Scenes from the Datasets selected for Overhead Detection with Transformers: RarePlanes, DOTA, and XView}
  \label{fig:detection:rareplanes}
\end{figure}

While convolutional-based deep neural networks have shown excellent performance in CV applications, with the recent publication of Visual Transformers, we have again seen a leap in computational vision capabilities.
As such, it is imperative to understand how various transformer-based neural networks perform on satellite imagery.
While transformers have shown high levels of performance in natural language processing and CV applications, they have yet to be compared on a large scale to modern RS data.
This comparison is especially necessary because oftentimes, algorithms and methodologies applied in ground photo CV applications do not effectively translate to RS applications, as the unique characteristics of RS imagery present challenges to new techniques.
Additionally, transfer learning has been shown to provide an incredible benefit to deep neural models (see \cite{CGINets}), however, most pretrained weights easily available are trained on ground photo datasets like ImageNet \cite{krizhevsky2017imagenet} and COCO \cite{coco}, which while robust, are not necessarily the most applicable dataset for transfer learning in the RS domain.
By training models on popular convolutional and transformer-based models on overhead benchmark datasets, we not only enable a comparison of their performance in the RS domain but also simultaneously generate weights that can be used for better transfer learning for overhead vision applications.

Meanwhile, to effectively evaluate transformer architectures on overhead imagery and generate pretrained weights for transformer architectures in the RS domain, an incredible amount of compute is also required.
Research has shown that transformer architectures require more training time and data to effectively train than their convolutional siblings.
To effectively evaluate transformers in the RS space, we must train a breadth of architectures utilizing differing feature extraction and detection techniques.
Further, in order to derive conclusions about any single architecture’s performance characteristics, we need to utilize multiple datasets of varying sizes.
Combining these factors necessitates thousands of hours of compute on powerful hardware in order to generate the number of trained models required to make conclusions on the applicability of transformers on overhead imagery.
This level of compute is available in Nautilus with its 1,300 NVIDIA GPUs, thousands of CPU cores, and terabytes of available memory.
Finally, orchestration and repeatability are paramount to ensuring sound scientific results, as well as scaling research compute.
Technologies such as Kubernetes and Docker enable this reproducibility and orchestrate the software needed to train and test such a vast number of models.

For this research, we compare ten distinct bounding-box detection and localization algorithms, of which seven were published since 2020, and all ten since 2015.
The performance of four transformer-based architectures is compared with six convolutional networks on three state-of-the-art open-source high-resolution RS imagery datasets ranging in size and complexity.
Descriptions of the networks utilized for this research as well as the datasets on which these networks are trained are described below:

\subsubsection{Two-Stage Convolutional Networks}
Two-stage convolutional-based methods perform region proposal as an initial feature processing step and then utilize these learned region proposals to perform bounding box localization.
The most popular method for two-stage detection is the Faster R-CNN \cite{faster_rcnn}, which utilizes a Region Proposal Network (RPN) to learn the set of region proposal from the most course learned feature map.
In this research, we utilize a Faster R-CNN as the detection methodology for several backbones, both convolution and transformer.
The convolutional backbone used for Faster R-CNN here is the ConvNeXt \cite{convnext} backbone.

\subsubsection{Single-Stage Convolutional Networks}
Unlike two-stage detectors, single stage networks learn the bounding box localization directly from the learned feature maps without needing prior region proposal.
These networks are generally less compute intensive and therefore more appropriate for real time applications.
The single-stage methods investigated in our comparison with transformers are Single Shot Detector (SSD) \cite{ssd}, RetinaNet \cite{retinanet}, Fully Convolutional One-Stage Object Detection (FCOS) \cite{FCOS}, and YOLOX \cite{yolox}.
The feature extraction architectures used for these networks are VGG16 \cite{vgg_net}, ResNeXt-101 \cite{resnext}, ResNet-101 \cite{resnet}, and YOLOX-XL, respectively.
\subsubsection{Visual Transformer Networks}
We train and evaluate three families of visual transformer networks to compare with the aforementioned convolutional models.
The first of these is the original Vision Transformer (ViT) \cite{vit}, which utilized transformer techniques from Natural Language Processing and modified them for use in CV.
The next set of transformer architectures utilized here is the DETR \cite{detr} and Deformable DETR \cite{deformable_detr} networks.
These networks, unlike ViT, are end-to-end detection networks utilizng only transformers and have shown promising result in traditional CV applications.
The final model used in this study is the SWIN Transformer, specifically SWIN-T \cite{swin}, which is an improvement on the original ViT in terms of both network performance and computational efficiency.
Since both SWIN-T and ViT are general-use feature extraction architectures rather than end-to-end detection networks, Faster R-CNN will be utilized as the bounding box localization for each of these models.

\subsubsection{Datasets}
Three open-source high-resolution remote sensing imagery (RSI) datasets are investigated in this study.
Utilizing multiple datasets in our investigation of transformer architectures is crucial for ensuring that any conclusions and analyses are not biased by a models' performance on a single dataset.
To that end, we select three visually and characteristically distinct datasets for this study, as can be seen in Figure~\ref{fig:detection:rareplanes}.
The first of these, the RarePlanes dataset \cite{rareplanes}, is a 7-class aircraft detection dataset containing just over 25,000 samples.
Next, the DOTA dataset \cite{dota}, is a general-use overhead RSI dataset containing around 250,000 objects belonging to 16 classes.
Finally, the XView dataset is a large-scale RSI dataset, containing over a million total objects in 60 classes.
Each of these datasets presents distinct challenges to our object detection models and should enable a robust comparison.
Note that for DOTA and XView, which contain large scenes, preprocessing is applied to chip the imagery and labels to 512x512 with overlap to enable more robust GPU acceleration on Nautilus.

\subsection{Burned Area Segmentation}
\label{sec:burned-area-methods}

Wildfires have become common all over the world.
Every year multiple fires burn hundreds of thousands of acres in %
North America and worldwide.
These fires can cover large areas and different land covers, which makes the task of precisely mapping the burned area (BA) an arduous one.
The exact mapping of the BA faces many obstacles,
such as difficulty to access the remote area where the fire took place,
the discrepancy between agencies and parties responsible for accomplishing this task, and human error.

Earth observation (EO) data opens an alternative possibility for field surveys and facilitates collaboration between different agencies, reducing discrepancy and missing data.

Sentinel-2 is one of the platforms used for acquiring EO data,
launched by the European Space Agency (ESA) through their program Copernicus.%
It offers rasters composed of 13 bands with the highest
of these medium-resolution sensors
at 10m and a revisit time of five days thanks to the twin satellites placed at 180 degrees from each other.
This makes Sentinel-2 an ideal, cost-effective sensor for BA mapping.

\begin{figure}[!t]
  \centering
  \includegraphics[width=0.5\linewidth]{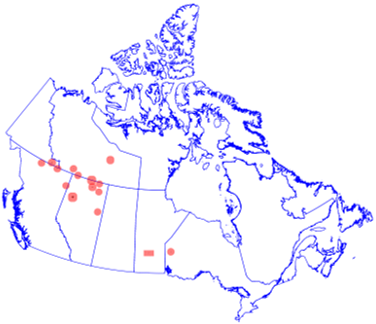} 
  \caption{Example of ground-truth polygons for Burned Area Mapping: red bounding boxes drawn around the burned area polygons}
  \label{fig:polygons}
\end{figure}

\begin{figure}[!t]
  \centering
  \includegraphics[width=0.5\linewidth]{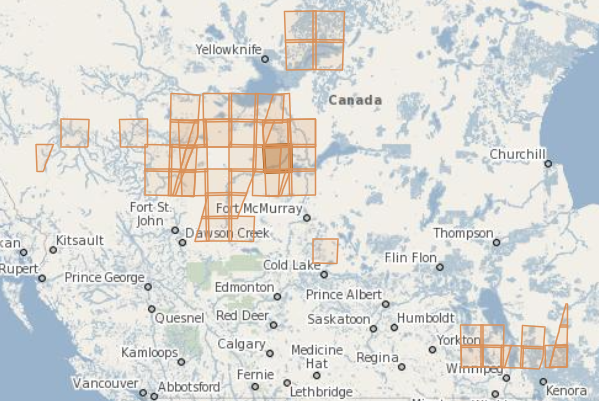}
  \caption{Examples of rasters to be downloaded for Burned Area Mapping. Each orange square represents a raster.} 
  \label{fig:rasters}
\end{figure}

A variety of approaches have been utilized for BA mapping.
Temporal studies that use spectral indices (SI) \cite{SIpaper} that combine different bands and compare their outcome before and after the fire were first developed.
These methods produced varying results depending on the bands used and land cover.
Machine learning techniques (ML), such as random forest, have also been used on Sentinel and Landsat data \cite{MLBA}.
Following recent trends in CV, deep learning has been employed for BA mapping, using approaches such as Siamese Self-Attention (SSA)\cite{DLBA1} and U-Net \cite{DLBA2}.
This problem is perfect for deep learning techniques, but not many have been employed so far.

DCNNs have been used for a variety of CV problems, from image classification to object detection.
Semantic segmentation is the natural next step in the list of CV tasks to be tackled by DCNNs where classification is performed at the pixel level.
The first attempt at deep learning semantic segmentation was done using a fully convolutional network (FCN) \cite{FCN}.
Then the encoder-decoder structures were used to build the U-Net \cite{UNet} and U-Net++ \cite{UNet++}.
In U-Net architectures, the image first goes through an encoder that is built using convolutional layers followed by pooling layers.
Next, the feature maps pass through a decoder where the features are upsampled using a transpose convolutional layer.
U-Net also possesses horizontal connections that connect layers in the encoder part with layers in the decoder part of the network.

DeepLab is another semantic segmentation network developed by Google, this network has seen many variants,
with the latest form DeepLabV3  \cite{Deeplab3} and DeepLabV3+\cite{Deeplab3+}.
DeepLab uses atrous convolutional layers to replace some of the pooling layers and to build a spatial pyramid to account for objects with varying scales.

We frame the BA mapping as a semantic segmentation problem for which we train DCNN models on satellite images. 
These satellite images require a large storage capability as well as computing power, a task for which the Nautilus HyperCluster is best suited. 
We developed a process to download and process these satellite images to create a dataset. 
We took advantage of the vast resources of Nautilus to train 144 models to determine the best hyperparameters for our problem. 
We then used these parameters to determine which of the four architectures produces the best results.

\subsubsection{Imagery}
\label{subsec: images}

The %
EO data we are using is Sentinel-2 L2A, bottom-of-atmosphere reflectance satellite imagery.
Each raster has 13 bands distributed over three different resolutions (10m, 20m, and 60m).
They are easy to obtain through their portal where the only requirement is to have an active account.
From the 13 bands, we select the Red, Green, and Blue bands at 10m resolution.
These three bands are then stacked together and normalized to create an RGB image.
While there are multiple different methods to normalize these bands,
we chose to normalize our bands using the 1st and 99th percentiles as the minimum and maximum values, respectively, for clamping and stretching the histogram.
In addition to the normalized image,
We also use the true-colored image (TCI) supplied with the raster bands.

\subsubsection{Burned area ground truth}
\label{subsec: Burned area ground truth}

The BA ground truth is in the form of Geo-tagged polygons.
These polygons are provided by the Canadian Wildland Fire Information System (CWFIS)\footnote{Burn Polygons: https://cwfis.cfs.nrcan.gc.ca/datamart/download/nbac?\\token=888fee58cba7b45d7a68ccbcc19dade1}.
We start from the ground truth polygons and begin by processing them to get the appropriate coordinates to download Sentinel-2L2A rasters, and for the creation of the semantic segmentation ground-truth mask. Figure~\ref{fig:polygons} shows the bounding boxes (BB) created around the burned area.
The area of each of these boxes is equal to or larger than the area of a single raster. We then use these coordinates to download the appropriate rasters, as shown in Figure~\ref{fig:rasters}.

\begin{figure}[!t]
  \centering
  \begin{tabular}{ccc}
    \includegraphics[width=0.2\linewidth]{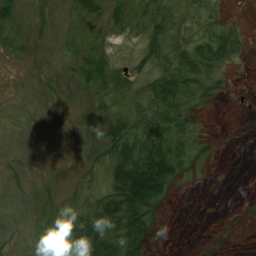} & \includegraphics[width=0.2\linewidth]{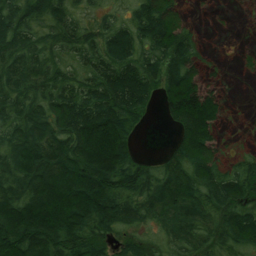}  & \includegraphics[width=0.2\linewidth]{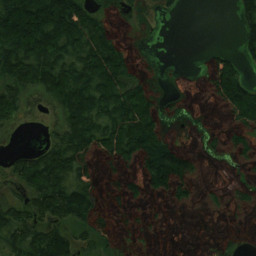}  \\
    \multicolumn{3}{c}{Images} \\
    \includegraphics[width=0.2\linewidth]{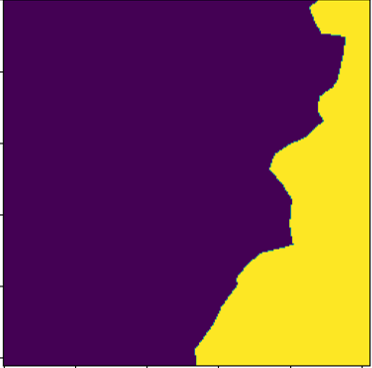} & \includegraphics[width=0.2\linewidth]{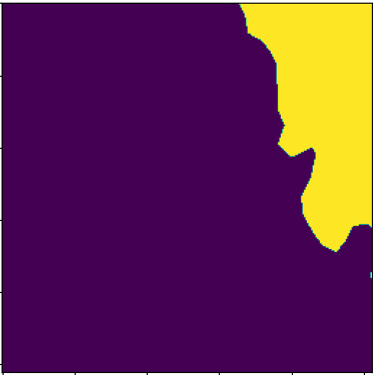} & \includegraphics[width=0.2\linewidth]{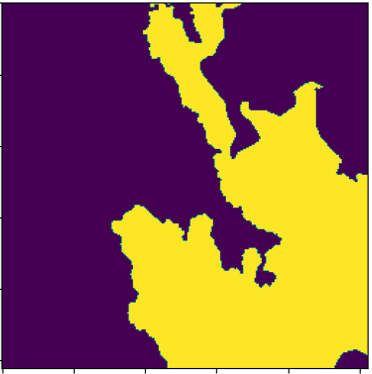} \\
    \multicolumn{3}{c}{Masks} \\
  \end{tabular}
  \caption{Example chips used for training Burned Area Mapping models}
  \label{fig:image_and_masks}
\end{figure}
\begin{table}[!t]
  \centering
  \caption{Number of jobs run and amount of data processed at each stage for Burned Area Mapping.}
  \label{tab:data-pipeline-statistics}
  \begin{tabular}{|l|c|c|c|c|c|}
  \hline
    \textbf{Phase} & \textbf{Download} & \textbf{Norm} & \textbf{Label} & \textbf{Chip} & \textbf{Total} \\
    \hline
    Jobs & 46 & 46 & 46 & 36 & 174\\
    \hline
    Data (GB) & 146 & 808.41 & 20.43 & 17.75 & 992.6 \\ 
    \hline
  \end{tabular}
\end{table}

Since the rasters we are using are from the year 2019, they are older than one year.
In the Copernicus hub, all images older than one year are automatically moved to Long Time Archive (LTA).
So each raster needs to be requested individually.
Once all rasters are online, they can be downloaded together in bulk.

A single Sentinel-2 L2A raster covers an area of 100km$\times$100km which, depending on the band resolution, could range from $10,000\times10,000$ pixels to $1830\times 1830$ pixels. There is a significant need for storage space that can only be satisfied by a cluster such as Nautilus. 
In addition, these rasters cover different areas at different periods of time, so each group of rasters needs to be downloaded separately and independently from each other.

Rasters are normalized as described in \ref{subsec: images}.
Next, We use the rasterize method in the Rasterio library along with the rasters' coordinates reference system and the ground truth polygons to create the semantic segmentation masks.
Lastly, we generate training chips from the larger scenes using a sliding window of size $256\times256$ pixels with 25\% overlap between every two consecutive chips.
We choose to include only the chips that contain both classes;
we set the threshold to select a chip to be at least 10\% of burned and unburned pixels, and we remove chips that do not meet this threshold.
Thus, the number of chips generated from each raster depends mainly on the size of the burned area.
We generated the input chips from both normalized rasters and TCIs, and we generated the corresponding mask chips from the created raster masks.
An example of the chips used in training is shown in Fig.~\ref{fig:image_and_masks}.

\subsubsection{Data Acquisition Scaling}

We treat each red box in Figure~\ref{fig:polygons} independently from each other, where we get a batch of rasters from each box. 
So we ran the entire data processing pipeline for each batch of rasters in parallel.
The coordinates of each of the red bounding boxes are used to acquire a batch of rasters, and each batch of rasters is downloaded using a Kubernetes job on Nautilus, with multiple jobs spun in parallel to download these rasters, enabling us to download our data in a very short period of time. 
We then use multiple jobs to process our rasters and normalize them, which allowed us to quickly process the entire dataset. 
Next, we generate a label mask for each raster, using the rasterize method from the Rasterio library. 
These masks are 10,000$\times$10,000 pixels each. 
Finally, we generate training chips from TCI images, normalized images, and label masks. 
Table~\ref{tab:data-pipeline-statistics} shows a breakdown of the number of jobs run at each stage as well as the amount of data processed.

We downloaded 144 rasters that generated a total of 7273 chips.
There were some redundant rasters that generated redundant chips. So we removed the redundant data.
Our final dataset has 63 rasters generating 5762 chips.
The number of chips generated by each raster varies from a few chips per raster to over 700 chips per raster.
Instead of blindly splitting our dataset 70\% for training, 20\% for validation, and 10\% for testing.
We chose to split our dataset by rasters.
We chose to use rasters that generated a large number of chips in our training and validation sets,
and use rasters with a few chips in our test set as this will make our test set more diverse.
The training set has 20 rasters in total generating 3928 chips representing 68\% of the total number of chips.
The validation set has three rasters in total generating 1120 chips representing 20\% of the total number of chips.
The test set has 40 rasters in total generating 714 chips representing 12\% of the total number of chips.
The dataset is staged on a persistent volume on Nautilus where the entire process is run.

\subsection{Deforestation Detection in the Brazilian Amazon}
\label{sec:deforestation-methods}

\begin{figure}[tbp]
\begin{center}
  \begin{tabular}{cc}
    \multicolumn{2}{c}{\includegraphics[width=0.5\linewidth]{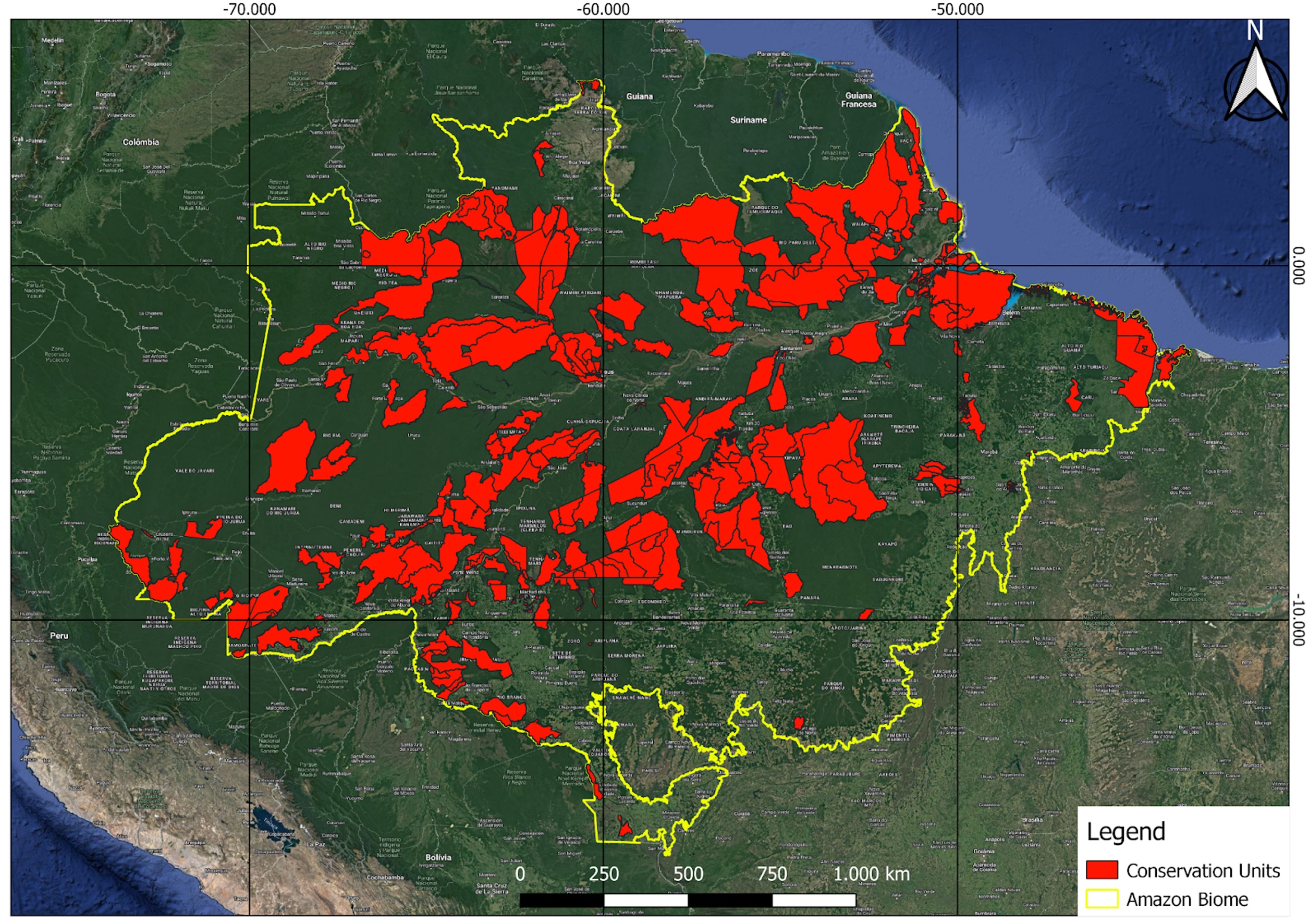}} \\
    \includegraphics[width=0.24\linewidth]{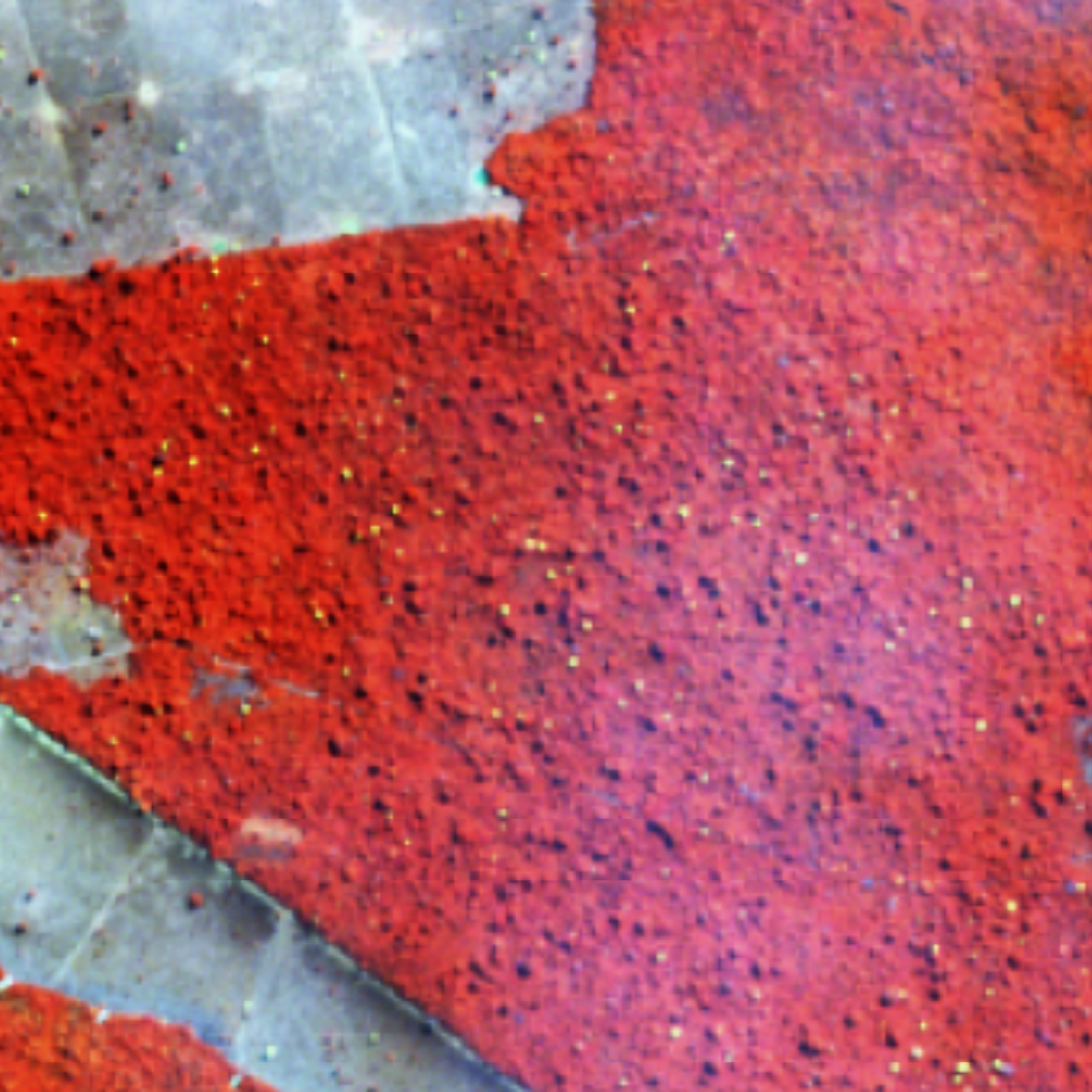} & \includegraphics[width=0.24\linewidth]{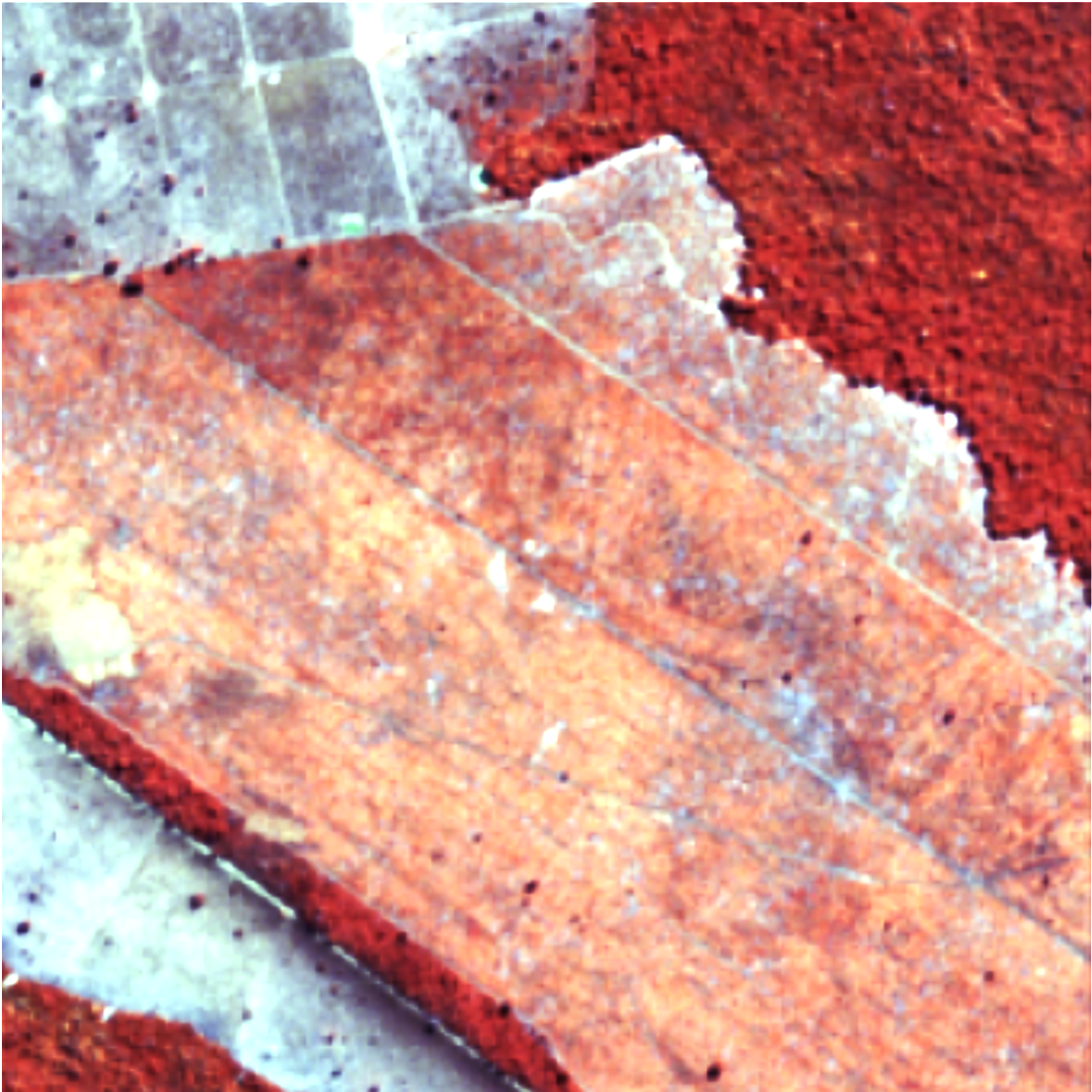} \\
  \end{tabular}
  \end{center}
  \caption{The Amazon Biome (top) is outlined by the yellow boundary and the conservation units are highlighted in red. Color-shifted infrared from chips for 2020 (left) and 2021 (right) highlight deforestation changes up close.}
  \label{fig:amazon_biome}
\end{figure}
Deforestation poses a severe threat to the natural ecosystem, causing a depletion of biodiversity, instability within ecosystems, and contributing to climate change.
As the largest rainforest in the world, the Brazilian Amazon plays a critical role in regulating climate and carbon levels.
Nevertheless, the rampant rate of deforestation has led to a host of issues, including increased greenhouse gas emissions, reduced carbon storage capacity, and a higher frequency of forest fires \cite{debemChangeDetectionDeforestation2020}.
To address the catastrophic implications of deforestation, it is crucial to implement effective policies that rely on accurate and timely data.
Deforestation detection (DD) is the primary means of obtaining this critical data to preserve the Brazilian Amazon.
Traditional DD methods such as map interpretation, field surveys, and ancillary data analysis are labor-intensive and time-consuming due to the vast size of the region (covering approximately 5.2 million km\textsuperscript{2} of land).
To combat this obstacle, RSI has emerged as a practical and cost-effective solution, providing consistent and repeatable data over large areas.
Among the available RSI sources, the European Space Agency's Sentinel-2 is an ideal open-access data source for DD, with its 10 m resolution and 5-day revisit rate.

In RSI change detection (CD) applications, quantitative analysis is done to identify surface changes from images captured at different timestamps.
With its attention mechanism, the transformer model exhibits promise in accommodating multi-temporal images, as it is scalable, captures long-range sequence features, and facilitates efficient parallel processing.
This study aims to investigate the effectiveness of transformer-based networks in detecting deforestation in the Brazilian Amazon.

\subsubsection{Data Acquisition}
We used Sentinel-2 Level-2A Surface Reflectance images with a maximum cloud cover percentage of 20\%.
We obtained ground truth polygons from the PRODES project datasets, provided by the Brazilian National Space Agency (INPE).
PRODES has monitored and quantified annual deforestation rates in the Brazilian Amazon rainforest since 1988, utilizing medium-resolution satellite imagery and a team of experts to visually interpret the data.
The PRODES data is accessible to everyone on the TerraBrasilis website  \footnote{Terrabrasilis Download: http://terrabrasilis.dpi.inpe.br/en/download-2/}.
Figure~\ref{fig:amazon_biome} (top) displays the polygons of the Amazon conservation units, as well as the boundary of the Amazon Biome.
We selected the top-5 conservation units with the highest deforested land areas for 2020 and 2021 (see Table~\ref{tab:top5conservation}).
To determine the appropriate date range for each selected area, we identified the earliest date $d_{1}$ and the latest date $d_{2}$ of all the images used to label that area in PRODES.
These dates fall in the Amazon’s dry season--typically from late June to early September-- when imagery is clearer with fewer clouds.
The selected date range for each area was then set to $[d_{1}-30\ days, d_{2}+30\ days]$.
This range ensures that the images used in our study were as close as possible to the labeling dates so significant deforestation is less likely to have developed during this interim period between image acquisition and labeling.

\begin{table}[btp]
  \centering
  \caption{The top five conservation units by size of deforestation in the Brazilian Amazon}
  \label{tab:top5conservation}
  \begin{tabular}{|p{1cm}|p{1.5cm}|p{2.3cm}|p{2cm}|}
    \hline
    \textbf{Area} & \textbf{Area}                    & \textbf{Deforestation}           & \textbf{Deforestation} \\
    \textbf{ID}   & \textbf{(km\textsuperscript{2})} & \textbf{(km\textsuperscript{2})} & \textbf{(\%)}          \\
    \hline
    59            & 16792                            & 528.33                           & 3.14                   \\
    89            & 13017                            & 214.59                           & 1.63                   \\
    55            & 1974                             & 108.92                           & 5.51                   \\
    291           & 20395                            & 114.70                           & 0.56                   \\
    165           & 9315                             & 91.31                            & 0.98                   \\
    \hline
  \end{tabular}
\end{table}
\subsubsection{Band Combinations}
Sentinel-2 images are composed of 13 bands at varying resolutions of 10~m, 20~m, and 60~m.
We selected the visible and near-infrared bands at 10~m resolution and Scene Classification Layer (SCL) at 20~m resolution.
The near-infrared band distinguishes vegetation from other features, while the SCL band acted as a mask to eliminate cloud or cloud-shadow areas.
We employed three band combinations, Color-shifted Infrared (NIR-R-G), Normalized Difference Vegetation Index (NDVI), and Enhanced Vegetation Index (EVI), to highlight vegetation spectral signatures.
Figure~\ref{fig:amazon_biome} (right) highlights the deforestation in two NIR-R-G images taken in 2020 and 2021.

\subsubsection{Dataset Creation}
To ensure a representative and unbiased dataset, we applied careful preprocessing procedures.
Linear normalization was first applied to facilitate image comparison on a consistent scale, with minimum and maximum values selected from the value histogram's 1\textsuperscript{st} and 99\textsuperscript{th} percentiles.
We applied filtering at both the raster and chip levels.
High-quality rasters were selected manually, while at the chip level, we removed single-class chips and those with less than 10\% ``change'' class area.
We used rotation augmentation at 90 and 180 degrees to increase dataset size and enhance model robustness.
At the end of the chip creation process a total of 7,734 pairs of chips in $256\times256$ size and 1,406 pairs of chips in $512\times512$ size were generated.
The dataset then was split into three subsets: (60\%, 20\%, 20\%) for training,  testing, and validation, respectively.

\begin{table*}[!ht]
  \centering
  \caption{Transformer Compute Statistics, including parameters optimized per network, as well as the per model GPU-time and GPU Memory across all 30 models.}
  \label{table:transformer-results}
  \begin{tabular}{|l|c|c|c|c|c|c|c|}
    \hline
    \multirow{2}{*}{\textbf{Model}} & \multirow{2}{*}{\textbf{Params (M)}} & \multicolumn{2}{c|}{\textbf{RarePlanes}} & \multicolumn{2}{c|}{\textbf{DOTA}} & \multicolumn{2}{c|}{\textbf{XView}}                                     \\

                           &                             & GPU-hours                       & VRAM (GB)                 & GPU-hours                  & VRAM (GB) & GPU-hours & VRAM (GB) \\
    \hline
    ConvNext               & 67.1                        & 18.4                            & 9.7                       & 63.4                       & 9.0       & 60.4      & 11.3      \\
    SSD                    & 36.0                        & 20.6                            & 7.4                       & 53.4                       & 8.6       & 57.1      & 8.9       \\
    RetinaNet              & 95.5                        & 56.3                            & 6.5                       & 56.5                       & 24.2      & 58.1      & 13.7      \\
    FCOS                   & 89.8                        & 24.5                            & 5.4                       & 59.8                       & 5.4       & 56.2      & 30.5      \\
    YOLOv3                 & 62.0                        & 50.9                            & 9.8                       & 56.7                       & 4.1       & 55.3      & 3.7       \\
    YOLOX                  & 99.1                        & 12.1                            & 21.1                      & 56.4                       & 42.2      & 56.9      & 40.1      \\
    \hline
    ViT                    & 97.7                        & 8.0                             & 12.6                      & 59.3                       & 24.7      & 60.3      & 25.1      \\
    DETR                   & 41.3                        & 11.0                            & 9.7                       & 58.1                       & 2.5       & 56.2      & 2.5       \\
    Deformable DETR        & 40.9                        & 16.7                            & 20.1                      & 59.1                       & 6.1       & 61.9      & 17.9      \\
    SWIN                   & 45.2                        & 22.7                            & 19.9                      & 57.7                       & 37.9      & 58.2      & 13.7      \\
    \hline
    \textbf{TOTAL}         & 674.6                       & 241.2                           & 122.2                     & 580.4                      & 164.7     & 580.6     & 167.4     \\
    \hline
  \end{tabular}
\end{table*}

\section{Experiments}
We now move to the experimental results and analysis for the three aforementioned scientific applications.
Each application utilizes its own training and testing methodologies and measures separate performance metrics based relevant to each scientific application.
Detailed descriptions of these results are given below.

\subsection{Overhead Object Detection with Transformers}
\begin{figure}[!t]
  \centering
  \includegraphics[width=0.4\linewidth]{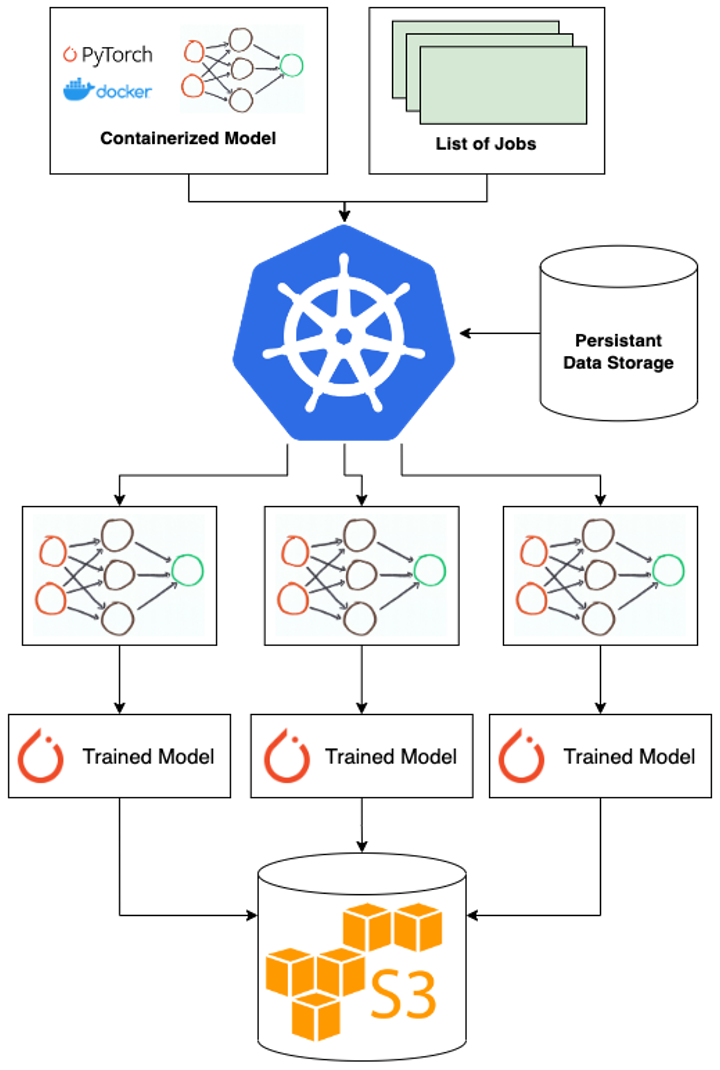}
  \caption{System Overview for Overhead Detection with Transformers}
  \label{fig:hurt_sys}
\end{figure}

Using Nautilus, we have trained and tested ten distinct models on three separate RSI datasets, generating 30 separate models.
Each of these ten architectures is trained on the three chosen datasets as described in Sect.~\ref{sec:object-detection-methods}.
All models are pretrained from COCO weights except for SWIN and ConvNext, which use ImageNet-1k weights, and all learning is performed with Stochastic Gradient Descent (SGD), except for SWIN and Deformable DETR, which are optimized using AdamW \cite{adamw}.
Hyperparameter configuration and dataset preprocessing vary by model and are set by mirroring the hyperparameters and image processing used in the generation of the pretrained weights, so as to enable the most effective transfer learning performance.
All pretrained weights used and all models trained come from the open source MMDetection framework \cite{mmdetection}.

Experiments are started in an automation fashion, in which containerized models are trained by Kubernetes jobs on the cluster as shown in Fig.~\ref{fig:hurt_sys}.
Data is staged in persistent volumes on the cluster prior to the beginning of training, and all models are copied to S3 cloud storage following training to ensure their later availability for evaluation.
Training and testing are performed utilizing environment variables that are updated using bash automation scripts that take in the set of models and datasets that are to be trained and launch each job on the Nautilus cluster in an automated fashion.
Four GPUs are utilized for each model's training, and the batch size is dynamically set based on available GPU memory, as the GPUs on Nautilus range from as little as the NVIDIA GTX 1080 (11 GB) to as high as the NVIDIA A100 (80GB).
In total, 30 models are trained in parallel on Nautilus, for a total of 9,000 epochs of training and over 2.02 billion learnable parameters optimized.
More than 137 TB of imagery are processed throughout the training processes and the total wall clock time of this compute is 2,142 hours.
Per-network learnable parameters along with per-model GPU-time and GPU memory are shown in Table~\ref{table:transformer-results}.

Reviewing the results of transformer model training statistics (Table~\ref{table:transformer-results}), we see that DOTA and XView required similar amounts of both GPU-time and GPU memory to generate the trained models, which is expected given the similar number of foreground images in each dataset.
RarePlanes, meanwhile, required less than half the GPU-hours to generate the ten trained models, but still required over 122 GB of VRAM.
Note as well the diversity in VRAM used by each model and dataset combination.
The increased VRAM used for YOLOX and FOCS enabled larger batch sizes, but this did not necessarily lead to lower training times, due in part to the incredibly high number of learnable parameters (over 90 million).

In terms of model detection performance, results ranged for each dataset and network.
However, the overall results indicated that SWIN and YOLOX were the best-performing networks, with AP50 scores exceeding 70\% on RarePlanes and 60\% on DOTA.
The results generated from this investigation will enable many insights into how transformers are able to perform on RSI and serve as a basis for performance expectations in real-world RS applications such as broad area search and other machine-assisted visual analytics applications.
Additionally, having these trained models on hand enables transfer learning for more novel RSI datasets in a variety of RS domains, as the performance benefits of transfer learning from relevant pretrained weights have been well documented in the CV space.

\begin{table}[tbp]%
  \begin{center}
    \caption{Comparison of the performance and training time of the four models: U-Net, U-Net++, DeepLabV3, and DeepLabV3+ using the best hyper-parameters}
    \label{table: four models}
    \begin{tabular}{ |l|c|c|c|c|c| }
      \hline
      & \multicolumn{4}{c|}{Class 1 (Burned Area)}                                                 & Time (s)    \\
      \hline
      \textbf{Model}                                  & \textbf{Prec (\%)}       & \textbf{Rec (\%)}       & $\mathbf{F_1}$          & \textbf{IoU}              & \textbf{Avg} \\
      \hline
      U-Net                                  
                                             & 83.03          & 80.88          & 0.82           & 0.694            & 22120             \\
      \hline
      U-Net++                               & 82.78          & 81.97          & 0.824          & 0.700             & 31716             \\
      \hline
      DeepLabV3                              & 83.71          & \textbf{83.78} & \textbf{0.837} & \textbf{0.720}   &  23731            \\

      \hline
      DeepLabV3+                             & \textbf{84.35} & 82.29          & 0.833          & 0.714            & 21607             \\
      \hline
    \end{tabular}
  \end{center}
\end{table}

\subsection{Burned Area Segmentation}

Training a DCNN requires the selection of crucial hyperparameters, which can be challenging for new problem domains, such as Burned Area Segmentation.
There exist two potential approaches to making these hyperparameter choices.
The first is that we can copy the training parameters from what may seem like a similar problem, in hopes that we generate a well-trained, generalizable model for our problem domain. 
The other approach is to train the model with different sets of these hyperparameters %
until we find the best combination. 
This approach is both time-consuming and compute-intensive, as 
the local machine we use may not have sufficient resources to train a model efficiently. %
Nautilus offers us a solution to this problem, with large amounts of compute resources that can be leveraged to train multiple large models utilizing Kubernetes, which then allows us to manage the training of multiple models in parallel. 
We utilize the resources available in Nautilus to search for the most appropriate combination of hyperparameters to train a DCNN for BA mapping. These parameters can then be used to train prominent DCNN architectures to see which network performs best.

We identified the set parameters that we want to investigate and select an appropriate set of values for each parameter.
We evaluated three different learning rates: $1\times10^{-3}$, $1\times10^{-4}$, and $1\times10^{-5}$
as well as three different batch sizes: 8,16 and 32.
We also examined the effect of weight initialization on our models, so we initialized our models once with weights pre-trained on the ImageNet dataset \cite{krizhevsky2017imagenet}, and once randomly.
Lastly, we examined two different optimizers \textit{Adam} and \textit{LAMB}.
In addition to the various hyperparameter values, we trained our models with two sets of data. The first set is from the chips we generated from the normalized RGB images and the other set is made up of chips generated from the TCI images provided by Sentinel-2 L2A.
Each combination of the values of these parameters constitutes an experiment, 72 in total.
We trained a U-Net model and a DeeplabV3 model with the ResNet-50 backbone for each experiment for 100 epochs.
In total 144 models were trained.
The training process is run on the same persistent volume where the data is staged.
We used a custom semantic segmentation framework based on Pytorch to train our models. This framework takes in a JSON configuration file where the specifics of each experiment are defined. We autogenerat these configuration files using the open-source Jinja2 library. 
For each experiment, we need two YAML files, one for training, and one for testing and evaluation. In total, we generate 288 YAML files. Instead of creating these files manually, we again relied upon Jinja2 library to autogenerate these files.
With the configuration files and the Kubernetes YAML files ready, we can proceed with the training of our models. 
Rather than manually submit these jobs, we opted to use a bash file to submit all jobs automatically. 
We were able to allocate 24GB of memory, four CPUs, and two GPUs for each model. 
We trained and evaluated 144 models in parallel for a total of 14400 epochs with 3.6 billion parameters optimized. We processed over five TB of imagery during the training process taking in a total of 518 hours of wall clock time.
Thanks to these experiments we determined that, the learning rate 1e-5, batch size 32, and optimizer \textit{LAMB} to be the best parameters to have smooth and stable decreasing training and validation losses.
We also found that the models initialized with the ImageNet dataset weight performed better than a randomly initialized model.

 We used these parameters to enhance the training of a broader set of models:
U\mbox{-}Net, U\mbox{-}Net++, DeepLabV3, and DeepLabV3+
We increased the number of epochs to 200 and introduced a scheduler such that the learning rate decreases by a factor of 0.5 every 50 epochs.
We were able to achieve the results shown in Table~\ref{table: four models}.

\begin{table*}[tbp]
  \centering
  \caption{Summary of Compute Performed on Nautilus}
  \label{table:compute-summary}
  \begin{tabular}{|l|c|c|c|c|c|c|c|}
    \hline
    \textbf{Scientific Application} & \textbf{Networks} & \textbf{Models} & \textbf{Parameters (M)} & \textbf{Imagery (GB)} & \textbf{Epochs} & \textbf{Time (h)} \\
    \hline
    Detection with Transformers     & 10                & 30              & 2024                    & 1370                  & 9000            & 2142              \\
    Burned Area Segmentation        & 4                  &      144           &       3600                  &    5175                   &         14400        &  518                 \\
    Deforestation Detection         &1                   &60                 &2460                         &31200                       &12000                 &1380                   \\
    \hline
    \textbf{TOTAL}                  &   15                &    234             &          8084               &    37745                  &      35200           &         4040          \\
    \hline
  \end{tabular}
\end{table*}
\subsection{Deforestation Detection in the Brazilian Amazon}
\begin{figure}[!t]
  \centerline{\includegraphics[width=0.9\linewidth]{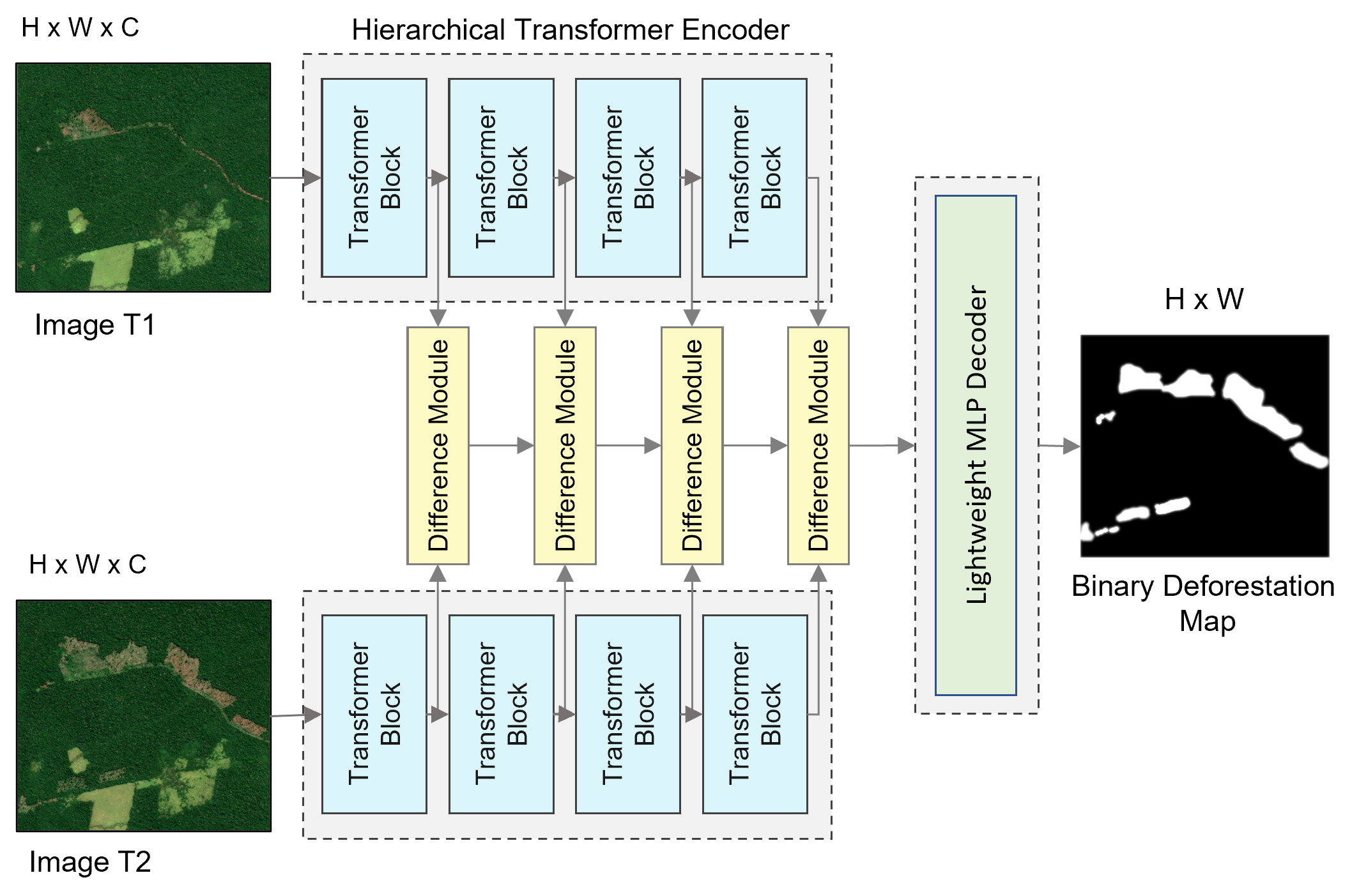}}
  \caption{A simplified overview of the ChangeFormer architecture, showing its three main components: a siamese hierarchical transformer encoder, four difference modules, and a lightweight MLP decoder.}
  \label{fig:change_former}
\end{figure}

ChangeFormer \cite{bandaraTransformerBasedSiameseNetwork2022}, a transformer-based network that is specifically designed for change detection, was used in this study. 
This network has demonstrated excellent performance in detecting both building and general changes in urban areas. 
Figure~\ref{fig:change_former} presents a simplified diagram of the ChangeFormer architecture. 
The model was randomly initialized and fine-tuned by adjusting its hyperparameters to optimize its performance.
We trained more than 60 model configurations for 200 epochs and evaluated their performance using overall and change-class-specific metrics such as F1 score, IoU score, precision, and recall.
Training entails processing over 112 million images of size 256$\times$256 and 8 million images of size 512$\times$512.
The cumulative wall-clock time for data pipeline and training is estimated to be 1380 hours.

We found that the 256$\times$256 chip size outperformed the 512$\times$512 chip size in all metrics, likely due to the availability of more training chips, with the top three results from the former surpassing the best result from the latter.
We further determined that the optimal combination for the ChangeFormer model was a learning rate of 0.0001, CE loss, AdamW optimizer, and the NIR-R-G band.
This achieved an overall accuracy of 94\%/92\% (256$\times$256/512$\times$512) with mIoU scores of 83\%/80\% and F1 scores of 90\%/88\%, respectively.
A previous study \cite{torresDeforestationDetectionFully2021} also applied deep learning for deforestation detection by comparing various fully convolutional network architectures.
Similar to our work, they used the PRODES dataset for ground truth, and they utilized both Sentinel 2 and Landsat-8 satellites imagery.
The best result for their study was achieved by FC-DenseNet with an F1-score of 70.7\%.
However, ChangeFormer model obtained at least 81\% F1-score for the change-class in both chip sizes.
Similar trends were observed in recall and precision, where we achieved 80\% and 86\% in our training process compared to 75.1\% and 78\% reported in the previous study.

\begin{figure}[!t]
  \centerline{\includegraphics[width=0.7\linewidth]{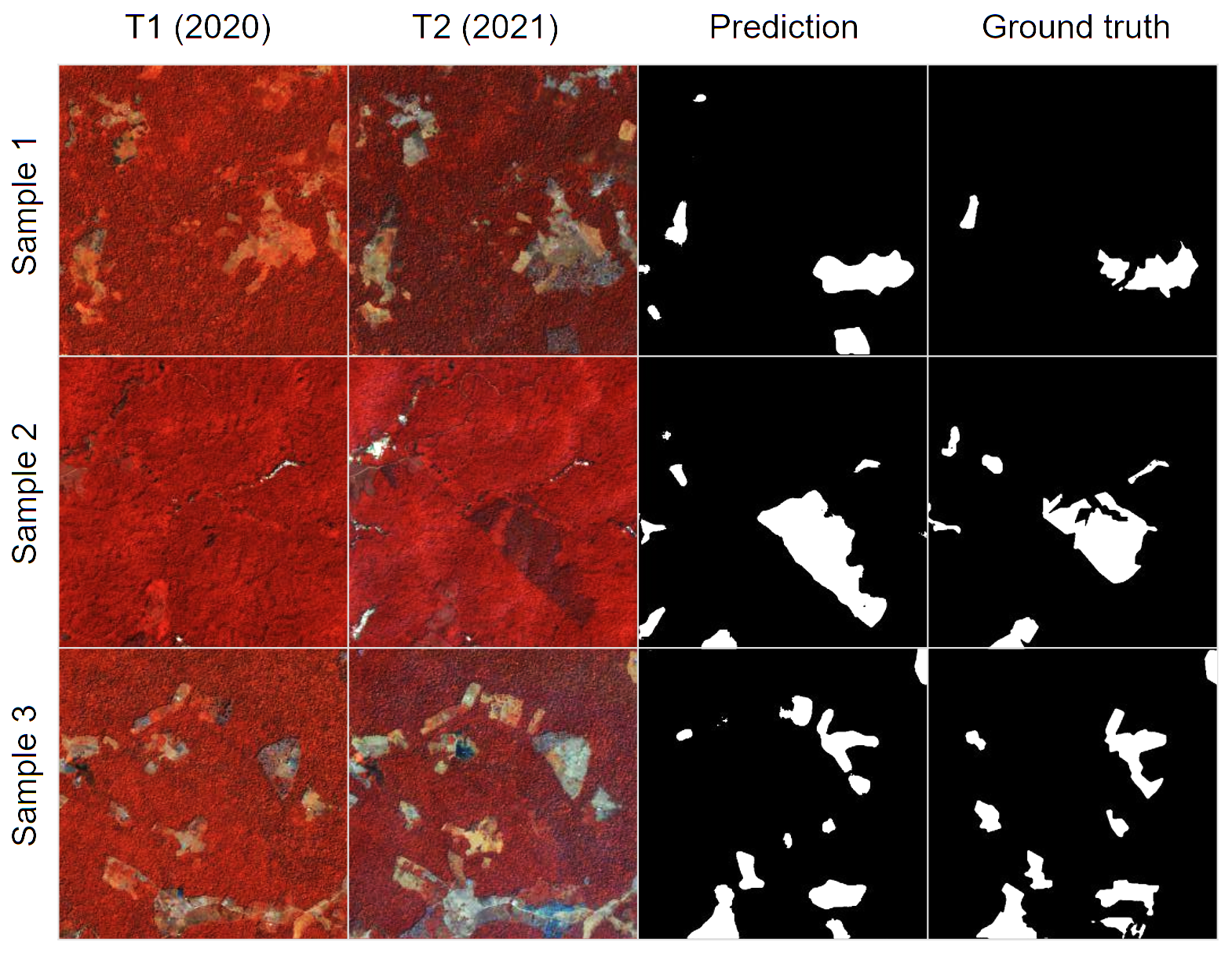}}
  \caption{Three image samples with corresponding predicted and ground truth deforestation maps. The first two columns show images from 2020 and 2021, respectively. The third and fourth columns display the predicted and ground truth maps, respectively.}
  \label{fig:predicted_ground_truth}
\end{figure}

\subsection{Scaling Overhead Vision with Nautilus}

In each of the above scientific applications, we have demonstrated a clear motivation for scaling deep learning research applications.
Effectively studying these problems requires an inordinate amount of compute that is not feasible for a single GPU-enabled server to complete in a reasonable amount of time.
To that end, we have scaled each of these research problems utilizing the Nautilus HyperCluster.
A summary of the compute performed can be found in Table~\ref{table:compute-summary}.

In total, over 4000 hours of compute are performed in parallel on Nautilus, the equivalent of over five and a half months if this compute were to be performed on a single server.
In performing this compute, over 37 TB of imagery is processed by the nodes on the cluster, and over 8 billion learnable parameters are optimized in the generation of over 230 models.
Scaling these research problems has enabled key takeaways and applications as well as opportunities for further research in each of these areas that would not have been possible without the scaling of the training methodologies to Nautilus.

\section{Conclusion}

Modern deep learning research is limited in many ways by the compute required to train and test models, even more so when applied at scale.
In this research, we have shown the benefits of utilizing the Nautilus HyperCluster to robustly automate and scale deep learning for high-resolution RS research.
We have demonstrated the benefits and ability to accelerate the training and testing of deep neural models for three separate deep learning applications in RS, including an exploratory investigation into deep neural transformers and the development of models capable of performing humanitarian aid: segmenting burned areas in North America and detecting deforestation in the Brazilian Amazon.
In our transformer investigation, we generated thirty sets of weights that can be used to enable more robust overhead RS applications.
Meanwhile, in burned area mapping, we were able to automate and explore the costly hyperparameter tuning required to apply deep neural models to new problem domains.
Finally, in applying DCNNs to deforestation detection, our trained models were able to outperform previously published competing methods by more than 10\%, due in part to the large grid of parameters that were able to be tested on the Nautilus cluster.

Future work for this research involves the development of more robust automation tools, perhaps utilizing the open-source Kubernetes Python API to build a Python library or application that can more easily and reliably manage jobs for training and testing deep learning models.
Additionally, future work will involve continuing to scale and improve model performance in each of these RS applications using deep learning on Nautilus.
Finally, utilizing Kubernetes to train models across multiple pods to train large models more effectively, regardless of the available GPU memory on a single pod would enable even more scalability, which may be necessary to effectively evaluate the incredibly large models being developed and released for RS research today.

\section*{Acknowledgement}

Computing resources for this research have been supported by the NSF National Research Platform and NSF OAC Award \#2322218 (GP-ENGINE).

\bibliographystyle{IEEEbib}
\bibliography{Nautilus_DeepLearning_HPC}
\end{document}